\documentclass{article}
\usepackage{color}
\usepackage{amsmath}
\usepackage{amssymb}
\usepackage[utf8]{inputenc}
\usepackage{url}
\usepackage{tikz}
\usetikzlibrary{shapes,arrows,positioning}
\tikzset{
    >=stealth',
    punkt/.style={
           rectangle,
           rounded corners,
           draw=black, very thick,
           text width=6.5em,
           minimum height=2em,
           text centered},
    pil/.style={
           ->,
           thick,
           shorten <=2pt,
           shorten >=2pt,}
}

\title{A probabilistic methodology \linebreak for multilabel classification}
\author{Alfonso E. Romero$^1$, Luis M. de Campos$^2$\\[0.5cm]
\small $^1$Centre for Systems and Synthetic Biology, and\\
\small   Department of Computer Science,\\
  \small Royal Holloway, University of London,\\
  \small Egham, TW20 0EX, United Kingdom\\
  \small \texttt{aeromero@cs.rhul.ac.uk}\\[0.5cm]
  \small $^2$Departamento de Ciencias de la Computaci\'on e Inteligencia
Artificial\\
  \small E.T.S.I. Inform\'atica y de Telecomunicaci\'on, Universidad de
Granada,\\
  \small 18071 – Granada, Spain\\
  \small \texttt{lci@decsai.ugr.es}
}
\date{}
\begin{document}

\maketitle
\thispagestyle{empty}
\begin{abstract}
Multilabel classification is a relatively recent subfield of machine learning.
Unlike to the classical approach, where instances are labeled with only one
category, in
multilabel classification, an arbitrary number of categories is chosen to label
an
instance. Due to the problem complexity (the solution is one among an
exponential number
of alternatives), a very common solution (the {\em binary} method) is frequently
used,
learning a binary classifier for every category, and combining them all
afterwards. The assumption 
taken in this solution is not realistic, and in this work we give examples 
where the decisions for all the labels are not taken independently, and thus, a 
supervised approach should learn those existing relationships among categories
to make 
a better classification. Therefore, we show here a generic methodology that can
improve 
the results obtained by a set of independent probabilistic binary classifiers,
by using a combination procedure with a 
classifier trained on the co-occurrences of the labels. We show an exhaustive
experimentation in 
three different standard corpora of labeled documents (Reuters-21578, Ohsumed-23
and RCV1), 
which present noticeable improvements in all of them, when using our
methodology, in three
probabilistic base classifiers.\\
{\bf Keywords:} {multilabel classification, label dependency, probabilistic
classifiers, text classification}

\end{abstract}

\section{Introduction}

In this work we present a novel solution for the multilabel categorization
problem. In this kind of problems,
a subset of categories (instead of just one) is assigned to each instance.
Multilabel classification problems 
arise, in a natural way, in information processing, concretely on the subfield
of automatic document categorization \cite{Sebastiani}.
 Due to their nature, an important number of text corpora are of a multilabel
kind. For example, news articles can often belong to 
more than one category (this is the case of the Reuters-21578 \cite{lewisr} and
the RCV1 \cite{lewisrcv1} collections, which are composed of 
articles from the Reuters agency). In other domains, multiple labels are
assigned as metadata which give a better description of the documents (this 
occurs, for example, in scientific papers and legal documents, which have
associated keywords from a controlled vocabulary, like the Mathematics Subject 
Classification or the MeSH Thesaurus in one case, and the Eurovoc thesaurus in
the other case \cite{ijar09}). On the other hand, multilabel 
instances occur very commonly in the Internet: in many blog applications, blog
posts for example, can be categorized with an arbitrary number of labels. 
Furthermore, in collaborative environments (like folksonomies, \cite{vanderwal})
where users can add tags, multilabel is an ordinary process. Due to this fact,
sometimes
the word ``tag'' or ``label'' is used instead of ``category'', they are all
synonyms.

More recently, multilabel classification has been useful for different domains
like, for instance, analysis of musical emotions \cite{trohidis,yang11}, 
scene or image categorization \cite{dimou,su11}, protein and gene function
prediction \cite{pandey,valentini11} or medical diagnosis \cite{taylor10}.

Although each instance has a given set of associated labels, and they are
assigned as a whole, the normal approach to solve this problem is just 
ignore this fact and concentrate in obtaining good solutions to the individual
binary problems (i.e., deciding for each label if it should be 
assigned to the instance or not). Here, we propose a solution which takes into
account the inter-category dependence, trying to find natural 
associations in order to improve the final categorization results.

The content of this paper is organized as follows: first of all (in Section
\ref{sec:multilabel}) we recall the well-known problem of multilabel 
supervised categorization, reviewing some previous works in this area (Section
\ref{sec:relatedwork}) together with a brief explanation 
(Section \ref{sec:semantic}) of what is the semantic of adding multiple labels
to an instance instead of one. Based on probabilistic foundations, our approach
is presented in Section \ref{sec:model}, which results in two different models.
An extensive experimentation with test collections coming from the text
categorization field will be carried out in Section \ref{sec:experiments} to
prove the validity of our proposal. 
Finally, in the light of the results previously obtained, several conclusions
and future works will be pointed out in Section \ref{sec:conclusions}.

\subsection{Multilabel supervised categorization}\label{sec:multilabel}

The problem of supervised multilabel classification (see, for instance
\cite{tsoumakas}) deals with supervised learning, 
where the associated labels can be a set of undetermined size. Formally, it can
be stated as follows: given a set of 
categories $\mathcal{C} = \{c_1, \dots, c_p\}$, an input instance space
$\mathcal{X}$, and a set of labeled data composed of 
instances and the set of assigned labels, $\{x_i, y_i\}_{i=1,\dots, n}$ where
$y_i \subset 2^\mathcal{C} \setminus \emptyset$, $x_i \in \mathcal{X}$, 
learning a multilabel classifier means inferring a function $f : \mathcal{X}
\longrightarrow 2^\mathcal{C}\setminus\emptyset$, in other words, a 
function able to assign non-empty subsets of labels to any unlabeled instance.

In order to cope with this exponential output space, the classical approach
consists in dividing the problem 
into $|\mathcal{C}|$ binary independent problems, and therefore learning
$|\mathcal{C}|$ binary 
classifiers $f_i : \mathcal{X} \longrightarrow \{c_i,\overline{c_i}\}$ (which
decide whether the category is assigned or not to the instance). Although 
this is a naive solution, it works reasonably well, and in multilabel
classification literature this can be used as a baseline. 
This approach is called the {\em binary relevance} method \cite{godbole,
tsoumakas, zhang}, and it is often criticized for ignoring the existing
correlations among labels.

Given the fact that many of the multilabel problems come from multi-tagged
collections (i.e., collections of objects which are
manually assigned a subset of categories of the whole set), it is very likely
that some of these tags are associated not only 
because of the content, but also due to the presence/absence of another tag. We
explain examples of this phenomenon in section 
\ref{sec:semantic}. Roughly speaking, the main motivation of this paper is to
look carefully at the results of the individual binary classifiers and
modify those results given a model previously trained on the label assignment
vectors, which explicitly captures relationships 
among categories and therefore improves classification results. As it is shown,
our approach utilizes a simple but powerful 
independence assumption which results in a model that is more complex than the
binary relevance method, but still is cheap enough.

\subsection{Related work}\label{sec:relatedwork}

There is more than one hundred references partially related with multilabel
categorization, most of them in the 
last years\footnote{See
\url{http://www.citeulike.org/group/7105/tag/multilabel}, and references
therein.}. At first sight, the 
two possible solutions to this problem basically consist of transforming it to a
single-label one, or adapting a learning procedure to work 
with these multiple labels at the same time. This taxonomy is given in
\cite{tsoumakas}, naming the former solutions {\em problem transformation
methods} 
and {\em algorithm adaptation} the latter ones. Because our contribution adapts
a learning procedure to multilabel learning, we review here only 
approaches of the second kind.

Almost all the typical classification algorithms have a multilabel version. For
example, an adaptation of the entropy formula of the C4.5 tree 
learning algorithm has been proposed in \cite{clare} for multilabel
classification. In the lazy algorithms field, variations of the $k$-NN 
algorithm have been presented for this kind of problems \cite{luo,zhang}. Also,
in \cite{cheng} a $k$-NN is combined with a logistic regression 
classifier (in a different way that we do) to cope with multiple labels. Of
course, variations on the SVM algorithm are shown in \cite{godbole}, 
where both intra-class dependencies and an improvement of the definition of
margin for multilabel classification are used to build a new model.

On the other hand, there are methods dealing with multilabel classification
within the probabilistic framework and therefore closer to our approach. 
In \cite{mccallum99}, a generative model is trained using training data, and
completed with the EM algorithm, and computed the most probable vector 
of categories that should be assigned to the document. A subset of Reuters-21578
is used for experimentation, and noticeable improvements are shown.

A generative model is also presented in \cite{ueda}. Here, the main assumption
is that words in documents belonging to several categories 
can be characterized as a mixture of characteristic words related to each of the
categories, being this assumption confirmed with experimentation.
 Both first (PMM1) and second (PMM2) order models are built, and learning
algorithms (using a MAP estimation) are proposed for both
 alternatives. Experiments are carried out with webpages gathered from the
\texttt{yahoo.com} server. Presented results are good, 
and improve other methods as SVM, naive Bayes and $k$-NN.

More recently \cite{dembe} proposed a novel method, called probabilistic
classifier chains (generalizing the classifier chains) which 
exploits label dependence, showing that the method outperforms others in terms
of loss functions. This is claimed via an extensive 
experimentation with artificial and real datasets.

\subsection{The semantic of assigning multiple labels}\label{sec:semantic}

In this section we try to enumerate three clear examples showing possible
reasons for a manual indexer
to assign multiple labels. Although we do not pretend to be exhaustive, we think
that
the three presented examples are common and may occur easily in multilabel
problems:

\begin{enumerate}
 \item {\em Mixture of topics}. An instance matches all the abstract description
of several categories.
This is the case, for example of a medical paper which deals with several topics
represented as MeSH keywords.

 \item {\em Contextualization}. Some tags are added in order to fix the context
in which other label is used.
For example, in scene classification, a picture of fishermen working in the
coast of Motril tagged with ``sea'' 
and ``people'' can be contextualized with ``town'' in order to distinguish it
from submarine photos.

 \item {\em Non overlapping labels}. Some subsets of labels do not admit
instances belonging to all of them. 
For example, in music classification, it is inconceivable to have songs tagged
with both ``baroque'' and ``reggae'', 
although other combinations as ``flamenco'' and ``jazz'' are possible.
\end{enumerate}

The only ``pure'' multilabel phenomenon is the first one. 
The second and the third denote that the occurrence of labels is not only based
on the content of the instance, but also in the 
occurrence (or not) of other labels. For the second case, a label is added to
contextualize two previously given labels. That is to say, the occurrence of 
a certain subset of labels increases the likelihood of other being added. On the
other hand, in the third case, 
a song labeled with ``baroque'' (at a certain degree) 
and ``reggae'' (in a lower degree) may be detected by a classifier as an
``anomaly'', and be labeled only with ``baroque'' 
(given that the system has 
previously learned that the label ``baroque'' gives information about the low
likelihood of also using the label ``reggae'' given that 
``baroque'' is being used).

Although the first phenomenon can be initially captured by different binary
classifiers, the second and the third can be tackled by looking at the labels of
a training set\footnote{Or adding expert knowledge with explicit relations among
the labels, like a hierarchy.}, and 
not only to the content of the instances. Therefore, our contribution will be to
state that the final labeling of an instance, in a certain category, will be a
combination of the result of the binary classifier with the evidence given in
the other categories by the other binary classifiers, taking into 
account the existing relationships among categories captured in the training
set. In fact, this issue of label dependence has been recently shown to be
crucial in multilabel learning \cite{dembe2}. 
All of this will be modeled in a probabilistic framework, which will give us a
rich language to describe this procedure.

\section{A probabilistic model for multilabel classification}\label{sec:model}


Let $x_i \in \mathcal{X}$ be an instance. Suppose we have a set of $p$
categories $\mathcal{C} = \{c_1, \dots, c_p\}$. For every category $c_j$, we
define a binary random variable $C_j = \{c_j, \overline{c_j}\}$. Let us assume
that we have $p$ probabilistic binary classifiers, based only on the content of
the instances (the features describing them). 
Thus, the conditional probability $p_j(c_j|x_i)$ represents the probability that
the instance $x_i$ is labeled with $c_j$ (the subindex $j$ indicates 
that this distribution is different and independent for every category).
Assuming we have a perfect knowledge of the underlying probability 
distribution, these classifiers define $p$ labeling rules as follows (Bayes
optimal classifier): ``classify $x_i$ as $c_j$ if $p_j(c_j|x_i) >
p_j(\overline{c_j}|x_i)$'' or, alternatively ``classify $x_i$ as $c_j$ if
$p_j(c_j|x_i) > 0.5$''.

For every category $c_j$, we shall also define a random vector of binary
variables ${\bf L_j} = (l_{j1}, \dots, l_{j p-1})$, which represents the
labeling of an instance with the other $p-1$ classifiers. We shall note for
$\mathbf{l_j}$ a particular value of the vector $\mathbf{L_j}$ (that is to say,
a label for the instance in all the categories except the $j$-th one). Thus,
every component $l_{jk}$ of the vector $\mathbf{l_j}$ is binary, and corresponds
to the variable $C_k$ if $k < j$ and to $C_{k+1}$ otherwise. 

Our aim is to give a model which describes the probability of a class given
knowledge of both the content of the instance, $x_i$, and the labeling of the
other categories ($\mathbf{l_j}$). In other words, an expression for the
probability $p_j(c_j|x_i,\mathbf{l_j})$.

\subsection{The proposed model}

Our model will start making a reasonable simplifying assumption ({\em general
naive Bayes assumption}\footnote{The term {\em general} is used here in a sense
of analogy with the ``classic'' naive Bayes assumption (given the true category
of an instance, the joint probability of its features factorizes as a product of
independent probability distributions). The previous labelings of the document
along with its content can be considered as two ``general'' features, and this
assumption means that we apply it to that set of two features.}): given the true
value of a category $j$, the events of finding a 
certain instance $x_i$ and a certain labeling on the other categories for this
instance, $\mathbf{l_j}$, are independent, that is

\begin{equation}
p_j(x_i, \mathbf{l_j}|c_j) = p_j(x_i|c_j) \, \, p_j(\mathbf{l_j}|c_j), \quad
\forall j \in \{1, \dots, p\}\,.
\label{nb}
\end{equation}

>From the point of view of a certain category $c_j$, if its value is known, this
equation assumes that the values of the 
other associated categories are probabilistically independent of the ``content''
of the instance $x_i$. While this assumption
might seem a bit unrealistic, it is basically the same which is performed in the
naive Bayes classifier (this is
why is called general naive Bayes). We shall show later that, as in the case of
the naive Bayes classifier, the assumption 
is neither intuitive nor much realistic but can result in very good
classification performance. Nevertheless, it should be stressed that our
assumption does not mean independence between labels at all, but independence
between labels and content given other label. As clearly expressed by the term
$p_j(\mathbf{l_j}|c_j)$ in eq. (\ref{nb}), our aim is to explicitly model a
clear dependence between each label $c_j$ and the other labels represented in
$\mathbf{l_j}$.

Then, using Bayes' theorem with this assumption, we compute the desired
probability:

\begin{eqnarray*}
p_j(c_j|x_i,\mathbf{l_j})&\, =\,& \, \frac{p_j(x_i,\mathbf{l_j}|c_j) \, p_j(c_j)}{p_j(x_i,\mathbf{l_j})}\, \\[0.5cm]
&\,= \,& \frac{p_j(x_i |c_j) \, p_j(\mathbf{l_j}|c_j) \,p_j(c_j)}{p_j(x_i,\mathbf{l_j})}\\[0.5cm]
 &\,=\,& \left(\frac{p_j(x_i)\, p_j(\mathbf{l_j})}{p_j(x_i,\mathbf{l_j})}\right)\left(\frac{p_j(c_j|x_i)\, p_j(c_j|\mathbf{l_j})}{p_j(c_j)}\right).
\end{eqnarray*}

The first term is a proportionality factor which does not depend on the
category. Therefore we get the expression,

$$
 p_j(c_j|x_i,\mathbf{l_j}) \propto
\frac{p_j(c_j|x_i)\,p_j(c_j|\mathbf{l_j})}{p_j(c_j)},
$$

\noindent which leads us to the final formula:

\begin{small}
\begin{equation}
\label{eq:probMultiLabel}
p_j(c_j|x_i,\mathbf{l_j}) = \frac{p_j(c_j|x_i)\,p_j(c_j|\mathbf{l_j}) /
p_j(c_j)}{p_j(c_j|x_i)\,p_j(c_j|\mathbf{l_j}) / p_j(c_j)\,\, + \,\,
p_j(\overline{c_j}|x_i)\,p_j(\overline{c_j}|\mathbf{l_j}) /
p_j(\overline{c_j})} 
\end{equation}
\end{small}

Taking into account that we are in binary classification, it holds that
$p_j(\overline{c_j}) = 1 - p_j(c_j)$, $p_j(\overline{c_j}|x_i)  = 1 -
p_j(c_j|x_i)$ and $p_j(\overline{c_j}|\mathbf{l_j})  = 1 -
p_j(c_j|\mathbf{l_j})$. The values $p_j(c_j|x_i)$ can be simply obtained with
any probabilistic binary classifier for the category $c_j$. Prior probabilities
$p_j(c_j)$ are estimated as the number of instances which belong to class $c_j$
over the total number of instances. Probabilities $p_j(c_j|\mathbf{l_j})$ can be
estimated, through a learning process, from the labels of the training data,
where every instance has as features some binary values telling if the instance
belongs or not to any of the $p-1$ categories (all categories except $j$). So,
in our model, we need to train $p$ binary classifiers from the content of the
instances, and $p$ binary classifiers from the labels assigned to these
instances.

A last point should be clarified in the model. Given an instance $x_i$ to
classify, we easily compute, for a category $c_j$ both the values $p_j(c_j)$ and
$p_j(c_j|x_i)$. However, to obtain the probability $p_j(c_j|\mathbf{l_j})$ we
would need to know the true assignments of the labels which are not $c_j$. In
our model, we shall make a second assumption, approximating $\mathbf{l_j}$ for
$\mathbf{\widehat{l_j}}$, which is computed as follows:

\begin{equation}
\label{eq:approximation}
\mathbf{\widehat{l_j}} = \big(\tau_k\left(\,p_k(c_k|x_i)\,\right)\big)_{k \in
\{1,\dots,p\} \setminus j}\,.
\end{equation}

Where $\tau_k$ is a threshold function ($\tau_k(z) = \overline{c_k}$ if $z <
0.5$, $\tau_k(z) = c_k$ otherwise). In other words, $$\widehat{l_{jk}} =
\arg\max_{\{c_k, \overline{c_k}\}} p_k(c|x_i).$$ Therefore
$p_j(c_j|\mathbf{l_j})$ will be approximated by
$p_j(c_j|\mathbf{\widehat{l_j}})$.

\subsection{An improved version of the model}

It should be noticed that the model summed up in eq. (\ref{eq:probMultiLabel})
does not really need the assumption that the random vectors $\mathbf{L_j}$ are
made of binary variables. In fact, the only required binary variables are the
$C_j$ ones. Thus, the same equation can be applied equally if the variables in
$\mathbf{L_j}$ are continuous. 

Therefore, we can rewrite an approximation of $\mathbf{l_j}$ different than the
one given in eq. (\ref{eq:approximation}) by removing the threshold function:

\begin{equation}
\label{eq:approximation2}
\mathbf{\widehat{l_j}} = \big(p_k(c_k|x_i)\big)_{k \in \{1,\dots,p\} \setminus
j}\,.
\end{equation}

This means that the components of the vector $\mathbf{\widehat{l_j}}$ are values
in $[0,1]$ which represent our degree of belief in these labels being assigned
to the instance. Note that, in order to use this extended version of the model,
we need a classifier to compute $p_j(c_j|\mathbf{l_j})$ which is capable of
dealing with continuous inputs (in the previous version we could assume the
inputs were binary, and so the classifier).

\subsection{The algorithm}

We present finally the proposed algorithm for multilabel classification. We
need, as input data, the prior
probabilities of each category ($p_j(c_j)$), $p$ content-only binary classifiers
(capable of providing an output 
$p_j(c_j|x_i)$), and $p$ classifiers which predict each category $c_j$ based on
the values of the labels different from 
$c_j$: $p_j(c_j|\mathbf{l_j})$.

\begin{enumerate}
 \item Given an instance $x_i$, we obtain the values $p_j(c_j|x_i)$ from $p$
binary probabilistic classifiers ($j=1,\dots,p$).
 \item For every category $c_j, j \in \{1,\dots,p\}$, 
  \begin{enumerate}
   \item Using $p_k(c_k|x_i)$, we compute ${\widehat{\mathbf{l_j}}}$ either by
eq. (\ref{eq:approximation}) or eq. (\ref{eq:approximation2}). This will be used
as an approximation of $\mathbf{l_j}$.
   \item We compute the likelihood of the category given the probabilities of
the other categories as $p_j(c_j|\mathbf{l_j})$.
   \item We compute the probability value $p_j(c_j|x_i, \mathbf{l_j})$,
following eq. (\ref{eq:probMultiLabel}).
  \end{enumerate}

 \item The scores of the instance $x_i$ in the set of categories $\mathcal{C}$
are the values $p_j(c_j|x_i, \mathbf{l_j})$. They can be thresholded,
optionally, if we are doing hard categorization.
 
\end{enumerate}

\tikzstyle{block} = [rectangle, draw, text width=10em, node distance=10em, text
centered, rounded corners, minimum height=3.5em]
\tikzstyle{cloud} = [draw, ellipse, node distance=10em,
    minimum height=2em, text width=2em, text centered]
\tikzstyle{invisible} = [ellipse, node distance=10em,
    minimum height=2em, text width=2em, text centered]

\begin{figure}[!hbt]
\centering
\begin{scriptsize}

\begin{tikzpicture}[node distance = 10em, auto, >=stealth] 
\node[cloud] (x) {$x$};
\node (AuxNode01) [text width=3.8cm, below=2 cm of x] {};

\node [block, left of=AuxNode01] (classifierJ) {classifier $p_j(c_j|x)$};
\node [block, right of=AuxNode01] (classifierALL) {classifier $p_k(c_k|x),
\linebreak k\in\{1,\dots,p\}\setminus j$};
\node [cloud, below=5 cm of AuxNode01] (eq1) {eq. ($\ref{eq:probMultiLabel}$) };
\node [block, below = 1cm of eq1] (result) { $\arg\max_{c_j} \linebreak
p_j(c_j|x,\widehat{\mathbf{l}})$ };
\node [block, below = 1cm of classifierALL] (argmax) { 
Use either eq. (\ref{eq:approximation}) or eq. (\ref{eq:approximation2})

};
\node [cloud, below = 1cm of argmax] (ele) { $\widehat{\mathbf{l}}$ };
\node [block, right of=eq1] (class) { classifier $p_j(c_j|\widehat{\mathbf{l}})$
};

\draw[->] (x) -- (classifierJ);
\draw[->] (x) -- (classifierALL);
\draw[->] (classifierALL) -- (argmax);
\draw[->] (classifierJ) |- (eq1);
\draw[->] (eq1) -- (result);
\draw[->] (argmax) -- (ele);
\draw[->] (ele) -- (class);
\draw[->] (class) -- (eq1);

\end{tikzpicture}
\end{scriptsize}
\caption{The general algorithm.}
\label{fig:diagram}

\end{figure}
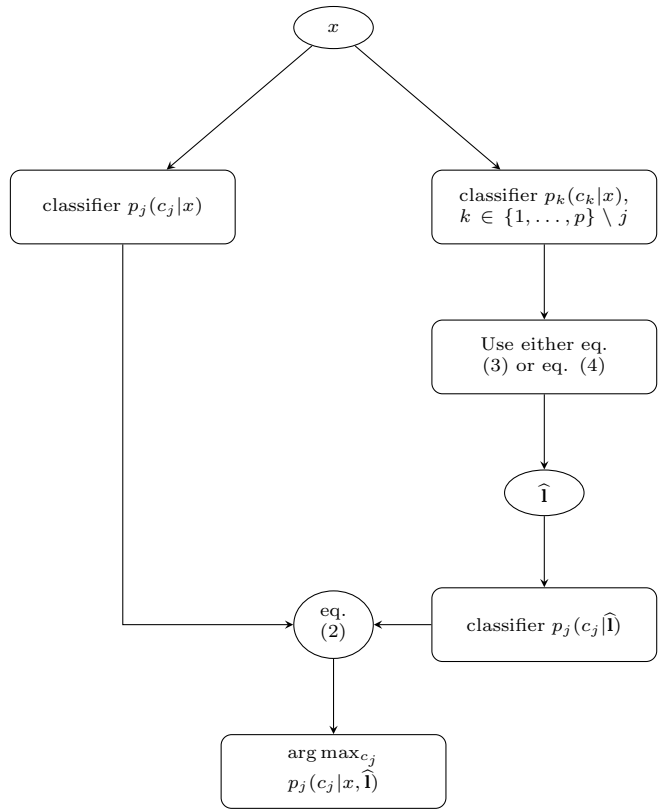

The whole process is summed up in Figure \ref{fig:diagram}.
Note that the computational load with respect to the binary relevance method is
very low. Initially, apart from the binary classifiers based on content
(which work on an input space containing, say, $m$ features), one label
classifier needs to be trained for each class. However, the label classifiers
usually will be very cheap to train, as they work on an input space possessing
$p-1$ 
features (and usually $p \ll m$). For the classification part, the binary
relevance method needs the example to be classified on
each of the $p$ binary classifiers. Once the scores of all the classifiers have
been obtained, $p$ extra classifications are needed
in the label classifiers (which should be much faster as the number of features
is low), and some additional computations needs to
be performed, summarized in eq. ($\ref{eq:probMultiLabel}$). As it can be seen,
neither much additional memory space, nor computational
power is needed to perform this approach. Therefore, its scalability should not
be an issue.

We have shown a new method to deal with relationships among labels, using
probabilistic classifiers.
 This is why it is presented as a ``methodology''. For a concrete problem, two
decisions should be made about which underlying models should be 
used: first for the content classifiers, and second for the label classifiers.
This choice relies heavily on the kind of problem selected, and is 
suitable of previous experimentation to find a working set of methods. Also one
may note that both choices are independent (so, with a set of $n_c$ 
probabilistic content classifiers and $n_l$ label classifiers, $n_c n_l$
combinations are possible, each one with two possibilities for 
computing the vectors $\mathbf{l_j}$).

\section{Experimentation}\label{sec:experiments}

We shall expose in this section an experimentation to test the validity of our
approach, where some 
combinations of classifiers will be selected following a certain criterion. Of
course, we do not aim being 
exhaustive and giving a long list of experiments. We are aware that many
different collections and combinations 
of classifiers could be selected, so we tried to make a good experimentation by
selecting a restricted but representative set of classifiers.

\subsection{Corpora}

Three different document categorization corpora have been used for
experimentation. We describe them now, together with the 
preprocessing procedure used to obtain the term vectors.

First, Reuters-21578 (see \cite{lewisr}, for instance) is a collection of
$21578$ news articles. We have used the most famous split 
(the one named ModApte), which divides the set of documents into a training and
a test set, and categories only assigned to documents in the 
test set are removed (then resulting only 90 of them).

The Ohsumed collection \cite{ohsu} is a set of $348566$ references from MEDLINE,
an online medical information database. For every record the assigned MeSH terms
(categories) are given. Because the number of categories of the MeSH thesaurus
is huge, it is often chosen a subset of 23 categories ({\em heart diseases}),
which are the root of some categories in a hierarchy. Documents which do not
belong to that subtree of categories, are discarded, and the resulting corpus is
called Ohsumed-23. This is the methodology followed in \cite{joachimsSVM}. 

Finally, the RCV1 corpus is a relatively more recent corpus (see
\cite{lewisrcv1}), also based on Reuters news stories. It contains many
documents ($806791$ for the final version, RCV1-v2), where the documents are
preprocessed, with stopwords removed, and terms already stemmed.  The number of
categories named ``topic codes''\footnote{Other two disjoint set of categories,
``industry codes'' and ``region codes'' are given for this corpus, but they are
not normally used for categorization.} is 103 (after the split). An standard
split is also provided, called LYRL2004, which gives a training set with over
$23000$ documents, and a test set with $781000$. We have removed two categories
which appear only
in the training set (reducing the number to 101).

In the first two cases, the stopwords list used consists of 571 stopwords of the
SMART retrieval system \cite{Salton3}. Also, the English stemming algorithm of
the Snowball package \cite{snowball} was applied to resulting words. In the
Reuters-21578 also XML marks were removed.

In order to reduce the size of the lexicon, terms occurring in less than three
documents were removed in Reuters-21578 and Ohsumed-23 (following the guidelines
of 
\cite{joachimsSVM}), and in less than five documents in RCV1, as done in
\cite{lewisrcv1}. 
All document preprocessing stage was made with DauroLab \cite{daurolab}.

\subsection{Evaluation measures}\label{sec:eval}

In this subsection we discuss the different evaluation measures we have used
for our experimentation. Following the taxonomy considered in \cite{pr12,miningmulti-label}, 
we have selected {\em label-based} measures, {\em example-based} and 
{\em ranking measures}. We shall present the chosen measures along with a
brief explanation of them, and a discussion of which performance aspects are
considered for each one.

\subsubsection{Example-based measures}

They are measures specifically designed for multilabel problems. That is, they
try to account for the performance of the multilabel task itself (the assignment
or not of a certain set of labels to one example). Of those available, we have
selected two: {\em Hamming loss} and {\em subset 0/1 loss}. Note that, as both 
measures are losses, the lower the value they have, the better the
classification
is.

Hamming loss computes the normalized Hamming distance
between the set of assigned labels and the set of predicted ones. That is, for a
certain
instance $x_i$, being $f(x_i)$ the set of predicted labels, and $y_i$ the set of
true labels,
the Hamming loss will be equal to:

$$\mathrm{HammingLoss} = \frac{y_i \, \Delta \, f(x_i)}{p},$$

\noindent where $\Delta$ stands for the symmetric difference of two sets and $p$
is the number of 
labels. The measure ranges 
between 0 and 1 and will be averaged through all the elements of the test set to
obtain its final value.

Subset 0/1 loss, in contrast, is a generalization of the 0/1 loss for multilabel
problems where
the set of labels is considered as a whole. For the previous case, the loss
function is equal to:

$$\mathrm{SubsetLoss} = I\left(y_i \ne f(x_i)\right).$$

In the previous equation, $I\left(y_i \ne f(x_i)\right) = 0$ if and only if $y_i
= f(x_i)$, and 1 otherwise. 
That is, there is no loss if the set of predicted labels equals the set of true
labels. Again, this measure
will be averaged through all the test set.

\subsubsection{Label-based measures}

These are classical measures for binary classification problems. We have made in
all cases hard 
categorization (assigning or not every label) and
then, we have used suitable measures for this task. Being defined for every
category $c_j$, the precision and recall ($\pi_j$ and $\rho_j$, respectively)
are:

$$\pi_j = \frac{TP_j}{TP_j + FP_j}, \quad \rho_j = \frac{TP_j}{TP_j + FN_j},$$

\noindent where $TP_j$, $FP_j$ and $FN_j$ stand for ``true positives'', ``false
positives'' and ``false negatives'' of the $j$-th category. We have then
selected the $F_1${\em-measure} adapted for categorization (the harmonic mean
between precision and recall), in its macro and micro averaged versions (denoted
by $MF_1$ and $\mu F_1$, respectively). See \cite{pr12,Sebastiani} for more details.

All the probabilistic classifiers have a natural threshold in $0.5$. We have not
made any threshold tuning because it was not the aim of this paper. Anyway, it
could be made independently, likely improving the results (see the discussion in
Section \ref{sec:conclusions}).

\subsubsection{Ranking measures}

We have implemented one ranking measure, the {\em one error}, which evaluates 
how many times the label ranked at the top is not in the set of relevant
labels of the instance. This measure is important for us, at it takes partially
into
account the ranking to perform the evaluation. For real-world use cases, in
multilabel
classification, where predicted labels are suggested to a human indexer using a
ranking procedure, it would be very important that the labels at the top of the list
were relevant.

\subsection{Basic probabilistic classifiers}

In order to have, first a baseline, and secondly a basic content classifier to
be used afterwards for our model, we have considered three different
classifiers, widely used in the literature. One which usually obtains discrete
results, and two with better results, all of them of a different nature. For the
first case, we selected the multinomial naive Bayes, in its binary version
\cite{mccallum}. For the second case, a $k$-NN classifier has been used,
normalizing the output in order to obtain a value in $[0,1]$ which can be
interpreted as a probability. Finally, a linear SVM, with probabilistic
output\footnote{The one implemented in LibSVM.}, as performed by Platt's
algorithm \cite{platt} was chosen. We recall that the flexibility of this
procedure is very high because any probabilistic classifier could be used
instead of these three.

For the $k$-NN classifier, the best performing value of $k$ was selected, based
on previous experimentation existing in other works. Thus, we chose $k=30$ for
Reuters and $k=45$ for Ohsumed-23 (as set in \cite{joachimsSVM}). For RCV1, a
value of $k=100$ was used \cite{lewisrcv1}. Also, following those references, we
have performed feature selection in Reuters (1000 features selected by
information gain, using the ``sum'' combination \cite{Sebastiani}), and in RCV1
(8000 features selected by $\chi^2$, using the ``max'' combination). No feature
selection was performed in Ohsumed-23 as noted in \cite{joachimsSVM}.

The implementation used for the algorithms was that contained in the DauroLab
\cite{daurolab} package, except for the SVM where libsvm \cite{libsvm} was
chosen.

\subsection{Label classifiers}

Here we show the two alternatives that we have selected as label classifiers.

\subsubsection{Logistic regression}

On the first hand, and based on previous experimentations (not shown here), we
have 
chosen a logistic regression-based classifier to model the posterior probability
distribution of the category given the correct labels of all the other
categories, namely:

$$p_j(c_j|\mathbf{l_j}) = \frac{1}{1 + e^{-(w_0 + \mathbf{w}\cdot
\mathbf{l_j})}},$$

\noindent where $w_0$ is a real scalar and $\mathbf{w}$ a $p-1$ dimensional real
vector, and ``$\cdot$'' is 
the usual dot product. The parameters of this model are learned to maximize a
certain criterion (generally a maximum likelihood approach).

We have selected this classifier instead of other proposals for three reasons:
first of all, it can deal with real inputs. Second, it is a discriminative
method, which does not try to find the joint distribution of $C_j$ and
$\mathbf{L_j}$ (in which we are not interested). Finally, it is very simple,
fast to learn, and works reasonably well in almost all the environments.

In order to have accurate estimates, and because the dimensionality of the
problem $p$ is high, the method selected to 
learn the weights is a Bayesian logistic regression, concretely the one proposed
in \cite{genkin}, with Gaussian priors. 
The implementation used here is the one included with the Weka package, with the
default parameters (see \cite{witten} for details).

One of the flexibilities of any logistic regression classifier is that it can
model distributions with real inputs. As we have two approximations for the
input vector in this classifier, we can propose different models, which will be
called M1 (corresponding to eq. (\ref{eq:approximation})) and M2 (for eq.
(\ref{eq:approximation2})). It seems reasonable that, if the content classifiers
perform well, the model M2 will be more accurate than the model M1. We shall
discuss this point afterwards. 

\subsubsection{Linear Support Vector Machine}

As a complement to the previous one, we have added the results obtained by using
a linear support vector machine with probabilistic outputs (fitted using Platt's
algorithm \cite{platt}).
Thus, the input to the SVM will be the vector with the output values of the
$p-1$ binary classifiers
$p_k(c_k|x),  k\in\{1,\dots,p\}\setminus j$, and the real-valued outputs of the
linear classifier
will be therefore transformed to probabilities with the model learned in the
training data by the mentioned algorithm.

In order to show the validity of our approach we have used a linear kernel
(i.e., the simplest SVM) without any
further modification to the default parameters to those included in the
canonical implementation of the Weka \cite{witten}
SMO (Sequential Minimal Optimization) class. It is well known that linear SVMs tend to perform reasonably well in
classification tasks, although 
they are generally outperformed by their kernelized counterparts. We have
selected the simplest model to illustrate
that we can improve in some measures the baseline of the binary classifiers
without too much effort.

\subsection{Results}

\subsubsection{Experiments with label-based measures}

We present the results for Reuters-21578 (in Table \ref{table:reuters}),
Ohsumed-23 (in Table \ref{table:ohsumed23}) and 
RCV1 (in Table \ref{table:rcv1}). NB denotes the multinomial naive Bayes, $k$-NN
the $k$ nearest 
neighbors classifier, and SVM the linear support vector machine with
probabilistic outputs. The terms `BLR' or `SMO' refers to the label classifiers
used, if any (Bayesian Logistic Regression, and linear Support Vector Machine,
respectively).
We have used the term SMO to distinguish this label classifier from the SVM in
the content
classifier. 

A classifier $+$ M1 or M2 denotes our proposal with the binary or 
real inputs presented to the corresponding label classifier, as explained
before. Also, the results of the base classifier have been shown for comparison
purposes.

We give the micro and macro $F_1$ values ($\mu F_1$ and $MF_1$, respectively),
and the percentage of improvement (positive or negative) over the baseline,
denoted by $\Delta$.

In all the cases our proposal with the M2 version of the algorithm
improved at least one of both measures of the baseline (i.e. $\Delta > 0$), of
which,
most of them had the two measures improved. The results were noticeable, even
reaching a gain of 72\% in the case of the macro $F_1$ measure with respect to
the baseline (which is a remarkable achievement). On the other hand, the M1
version
presented not so systematic improvements. 

Moreover, we have run statistical significance tests for every couple of
classifiers (a basic probabilistic classifier, and our proposal, either with the
M1 or the M2 configuration). For the micro measure, a micro sign s-test was
performed, and for the macro measure, we chose the macro S-test. Both are
presented in \cite{yang99} and constitute the nowadays standard for comparing
this kind of experiments. Nevertheless it should be noticed that the s-test is
not specifically designed for the $F_1$ measure, taking into account both true
positives and true negatives (as shown in Section \ref{sec:eval}, $F_1$ only
considers true positives). Also, note that the macro S-test does not take into
account the amount of improvement, but the number of categories where the
measure is improved, leading sometimes to counter-intuitive results (with a
non-significant high improvement in macro due to the improvement in only very
few categories).

In the test the first system, $A$, will be the best performing one, in terms of
micro or macro $F_1$ measure, being $B$ the second. The null hypothesis is that
$A$ is similar to $B$. On the other hand, the alternative hypothesis says that
$A$ is better than $B$.

If the p-value is less than $0.01$, we show in the tables the sign $\ll$ (resp.
$\gg$) to denote that our baseline plus M1 or M2 is significantly better (resp.
worse) than the baseline alone. The signs $<$ and $>$ are used to indicate the
same fact if the p-value is between $0.01$ and $0.05$. With the sign $\sim$ we
point out no significant difference between the two systems. Table
\ref{table:resumen} displays a summary of the results, showing the number of
times that each one of our models is better, significantly better, worse and
significantly worse than the baseline.

\begin{table}
\begin{center}
\begin{footnotesize}
\begin{tabular}{|c|ccc|ccc|}\hline
Model  & $\mu F_1$ & $\Delta$ &  s-test  & $MF_1$ & $\Delta$ & S-test\\\hline
NB &  0.61766 &  -- & -- & 0.25145  & -- & --\\
NB + BLR + M1 & 0.65697 & +6.36\% & $\ll$ & 0.29183 & +16.06\% & $\ll$ \\
NB + BLR + M2 &  0.65830 & +6.58\% & $\ll$ & 0.29293 & +16.49\% & $\ll$ \\
NB + SMO + M1 &  0.65644 & +6.28\% & $\ll$ & 0.29357 & +16.75\% & $\ll$ \\
NB + SMO + M2 &  0.66032 & +6.91\% & $\ll$ & 0.29650 & +17.92\% & $\ll$ \\
 &  & & & & &\\ 
k-NN &  0.78875			&  -- & -- & 0.27886  & -- & --\\
k-NN + BLR + M1 & 0.65255 & -17.27\% & $\gg$ & 0.34035 &+22.05\% & $\sim$ \\
k-NN + BLR + M2 & 0.78022 & -1.08\% & $\gg$ & 0.40246 & +44.32\% & $\ll$ \\
k-NN + SMO + M1 & 0.62175 & -21.17\% & $\gg$ & 0.26160 & -6.19\% & $\sim$ \\
k-NN + SMO + M2 & 0.79230 & +0.45\% & $\sim$ & 0.34944 & +25.31\%& $\ll$\\
 &  & & & & &\\
SVM & 0.87102 & -- & -- & 0.48529 & -- & -- \\
SVM + BLR + M1 & 0.85032 & -2.38\% & $\gg$  & 0.54401 & +12.10\% & $<$ \\
SVM + BLR + M2 &  0.87970 & +1.00\% & $\sim$ &0.55982	& +15.36\%& $\ll$ \\
SVM + SMO + M1 &  0.71923 & -17.43\% & $\gg$ &0.42208	& -13.02\% & $\gg$\\
SVM + SMO + M2 & 0.86832 & -0.31\% & $\gg$ & 0.49683 & +2.38\% & $\sim$ \\\hline
\end{tabular}
\caption{Label-based measures for Reuters-21578 corpus}
\label{table:reuters}
\end{footnotesize}
\end{center}
\end{table}

\begin{table}
 \begin{center}
  \begin{footnotesize}
 \begin{tabular}{|c|ccc|ccc|}\hline
Model  & $\mu F_1$ & $\Delta$ &  s-test  & $MF_1$ & $\Delta$ & S-test\\\hline
NB & 0.56166 & -- & -- & 0.49985 & -- & --\\
NB + BLR + M1 & 0.57789 & +2.89\% & $\ll$ & 0.52204 & +4.44\% & $\ll$ \\
NB + BLR + M2 & 0.57832 & +2.97\% & $\ll$ & 0.52208 & +4.45\% & $\ll$ \\
NB + SMO + M1 & 0.57635& +2.62\% & $\ll$ & 0.52099 & +4.23\% & $\ll$\\
NB + SMO + M2 & 0.57693 & +2.72\% & $\ll$ & 0.52136 & +4.32\% & $\ll$ \\
 &  & & & & &\\
k-NN &  0.43487 &  -- & -- & 0.29451 & -- & --\\
k-NN + BLR + M1 & 0.54352 & +24.98\% & $\sim$ & 0.50804 & +72.50\% & $\ll$ \\
k-NN + BLR + M2 & 0.47816 & +9.95\% & $\ll$ & 0.35945 & +22.05\% & $\ll$ \\
k-NN + SMO + M1 &  0.49761 & +14.43\% & $\sim$ & 0.41588 & +41.20\% & $\ll$\\
k-NN + SMO + M2 &  0.44502 & +2.33\% & $\sim$  & 0.31433 & +6.73\% &$\ll$\\
&  & & & & &\\
SVM & 0.64802 & -- & -- & 0.59596 & -- & -- \\
SVM + BLR + M1 & 0.66669 & +2.88\% & $\sim$ & 0.62482 & +4.84\% & $\ll$ \\
SVM + BLR + M2 & 0.65871 & +1.65\% & $\sim$ & 0.61175 & +2.65\% & $\ll$ \\
SVM + SMO + M1 &  0.65716 & +1.41\% & $\sim$ & 0.61258 & +2.79\% & $\ll$ \\
SVM + SMO + M2 &  0.65198 & +0.61\% & $<$ & 0.60393 & +1.34\% & $<$ \\\hline

\end{tabular}
\caption{Label-based measures for Ohsumed-23 corpus}
\label{table:ohsumed23}
\end{footnotesize}
\end{center}
\end{table}

\begin{table}
 \begin{center}
\begin{footnotesize}  
 \begin{tabular}{|c|ccc|ccc|}\hline
Model  & $\mu F_1$ & $\Delta$ &  s-test  & $MF_1$ & $\Delta$ & S-test\\\hline
NB & 0.61823 & -- & -- & 0.36539 & -- & -- \\
NB + BLR + M1 & 0.64282 & +3.98\% & $\ll$ & 0.39174 & +7.21\% & $\ll$ \\
NB + BLR + M2 & 0.64374 & +4.13\% & $\ll$ & 0.39245 & +7.41\% & $\ll$ \\
NB + SMO + M1 & 0.61358 & -0.75\% & $\gg$ & 0.37401 & +2.36\% & $\sim$ \\
NB + SMO + M2 & 0.60962 & -1.39\% & $\gg$ & 0.37287 & +2.05\% & $<$ \\

 &  & & & & &\\
k-NN &  0.68698 &  -- & -- & 0.27494  & -- & --\\
k-NN + BLR + M1 & 0.60402 & -12.08\% & $\gg$ & 0.40528 & +47.40\% & $\ll$ \\
k-NN + BLR + M2 & 0.72488 & +5.52\% & $\sim$ & 0.40404 & +46.96\% & $\ll$ \\
k-NN+ SMO + M1 & 0.51514 & -25.01\% & $\gg$ & 0.33489 & +21.80\% & $\ll$ \\
k-NN+ SMO + M2 & 0.57233 & -16.69\% & $\gg$ & 0.32626 & +18.67 & $\ll$ \\

 &  & & & & &\\
SVM & 0.80570 & -- & -- & 0.55609 & -- & -- \\
SVM + BLR + M1 & 0.78305 & -2.81\% & $\gg$ & 0.57704 & +3.77\% & $\sim$ \\
SVM + BLR + M2 & 0.80767 & +0.25\% & $\sim$ & 0.59334 & +6.70\% & $\ll$\\
SVM + SMO + M1 & 0.68165 & -15.40\% & $\gg$ & 0.54189 & -2.55\% & $\gg$ \\
SVM + SMO + M2 & 0.67950  & -15.55\% & $\gg$ & 0.56275 & +1.20\% &
$\sim$\\\hline

\end{tabular}
\caption{Label-based measures for RCV1 corpus}
\label{table:rcv1}
\end{footnotesize}
\end{center}
\end{table}

\begin{table}
 \begin{center}
\begin{tabular}{|c|c|c|}\hline
 & $\mu F_1$ & $MF_1$ \\\hline
BLR + M1 & 5/3/\textit{4}/\textit{4} & 9/7/\textit{0}/\textit{0} \\
BLR + M2 & 8/4/\textit{1}/\textit{1} & 9/9/\textit{0}/\textit{0} \\ 
SMO + M1& 4/2/\textit{5}/\textit{5} & 6/5/\textit{3}/\textit{2} \\
SMO + M2& 5/3/\textit{4}/\textit{4} & 9/7/\textit{0}/\textit{0} \\\hline
\end{tabular}
\caption{Number of times that the baseline classifier plus a label classifier
with either M1 or M2 is 
better/significantly better/\textit{worse}/\textit{significantly worse} than 
the corresponding baseline classifier alone, with respect to micro and macro
$F_1$}
\label{table:resumen}
\end{center}
\end{table}

Once shown the results, we may state that the methodology presented is useful
for improving classification results in a multilabel environment. Concretely,
in the presented tables, the following facts can be found:
\begin{itemize}
\item The use of this technique clearly improves the classification results of
the baseline. For the macro experiments, all the measures are improved from the
baselines. In the micro ones, also good improvements are found, especially for
the M2 version of the classifier and BLR, which improves the baseline in all but
one cases (Reuters with $k$-NN, where the loss is around $1\%$). The SMO
classifier went
a bit worse than the BLR, presenting in general smaller deltas, and worsening
all the
micro results in RCV1.

\item Comparing both approaches, the model M1 (binarized) performs, in general,
worse than the continuous (M2) version, regardless of the label classifier used.
This is due to the binarization procedure, which removes some information
represented in 
the granularity of the assignment that is well captured by the classifier (if
for a class $c_j$, the base classifier assigns to two instances the
probabilities 0.99 and 0.55, the binarized version will treat these two cases in
the same way, assuming that both examples are of class $c_j$; however the
continuous version will clearly distinguish them). 

\item In general, we can say that this methodology seems to benefit
classification of less populated categories (those with a very low prior) a lot,
without harming more frequent categories. This produces the fact that, in
general, the macro measures (where all categories weight the same) are heavily
lifted up, whereas micro measures are improved at a lesser degree. 

\item The improvement for the micro measure in the RCV1 corpus were quite small,
but it should be noted that the baseline value (0.80570) was very close to the
best result obtained by Lewis \cite{lewisrcv1} with SVM (0.816) and a
sophisticated threshold tuning algorithm (ScutFBR.1). Perhaps results better
than this high value are very difficult to get.
\end{itemize}

\subsubsection{Experiments with example-based and ranking measures}

We present the results for Reuters-21578 (in Table \ref{table:reuters_multi}),
Ohsumed-23 (in Table \ref{table:ohsumed23_multi}) and 
RCV1 (in Table \ref{table:rcv1_multi}). The rest of the notation is similar to
the one used in the previous tables. We have added the symbol $\bullet$ to the
values which are better or equal than the corresponding baseline, and $\circ$
to those which are worse.

\begin{table}
\begin{center}
\begin{tabular}{|c|ccc|}\hline
Model	&Hamming loss	& Subset 0/1 loss & One error\\\hline
NB	&0.01530 $\,\,\,$		&0.39020 $\,\,\,$	&0.85624 $\,\,\,$\\
NB + BLR + M1&	0.01291 $\bullet$	&0.37960 $\bullet$	&0.84929 $\bullet$\\
NB + BLR + M2&	0.01281	$\bullet$	&0.37827 $\bullet$	&0.84763 $\bullet$\\
NB + SMO + M1&	0.01330	$\bullet$	&0.37198 $\bullet$	&0.83107 $\bullet$\\
NB + SMO + M2&	0.01293	$\bullet$	&0.36734 $\bullet$	&0.83107 $\bullet$\\
&	&	&\\
KNN	&0.00512 $\,\,\,$		&0.26532 $\,\,\,$	&0.86651 $\,\,\,$\\
KNN + BLR + M1	&0.01250 $\circ$ &0.34316 $\circ$	&0.82544 $\bullet$\\
KNN + BLR + M2	&0.00600 $\circ$ &0.30176 $\circ$	&0.86585 $\bullet$\\
KNN + SMO + M1	&0.01401 $\circ$ &0.29911 $\circ$	&0.82113 $\bullet$\\
KNN + SMO + M2	&0.00567 $\circ$ &0.27791 $\circ$	&0.86154 $\bullet$\\
&	&	&\\						
SVM	&0.00334 $\,\,\,$		& 0.18980 $\,\,\,$	&0.94270 $\,\,\,$\\
SVM + BLR + M1	& 0.00427 $\circ$	& 0.20338 $\circ$	&0.92978 $\bullet$\\
SVM + BLR + M2	& 0.00331 $\bullet$	& 0.17390 $\bullet$	&0.93839 $\bullet$\\
SVM + SMO + M1	& 0.01162 $\circ$	& 0.19212 $\circ$	&0.89864 $\bullet$\\
SVM + SMO + M2 & 0.00709 $\circ$	& 0.20669 $\circ$	&0.88937 $\bullet$\\\hline
\end{tabular}
\caption{Example-based and ranking measures for Reuters}
\label{table:reuters_multi}
\end{center}
\end{table}

\begin{table}
\begin{center}
\begin{tabular}{|c|ccc|}\hline
Model	&Hamming loss	& Subset 0/1 loss & One error\\\hline
NB & 0.08318 $\,\,\,$&	0.83933 $\,\,\,$& 0.71595 $\,\,\,$\\
NB + BLR + M1 & 0.07687 $\bullet$ & 0.83593 $\bullet$ & 0.71477 $\bullet$\\
NB + BLR + M2 & 0.07664 $\bullet$ & 0.83462 $\bullet$ & 0.71516 $\bullet$\\
NB + SMO + M1 & 0.07668 $\bullet$ & 0.83698 $\bullet$ & 0.71556 $\bullet$\\
NB + SMO + M2 & 0.07651 $\bullet$ & 0.83567 $\bullet$ & 0.71595 $\bullet$\\
&	&	&\\		
KNN & 0.05494 $\,\,\,$ & 0.77823 $\,\,\,$ & 0.71687 $\,\,\,$\\
KNN + BLR + M1 & 0.07335 $\circ$ & 0.77692 $\bullet$ & 0.66479 $\bullet$\\
KNN + BLR + M2 & 0.05384 $\bullet$ & 0.75455 $\bullet$ & 0.71255 $\bullet$\\
KNN + SMO + M1 & 0.07308 $\circ$  & 0.75795 $\bullet$ & 0.63849 $\bullet$\\
KNN + SMO + M2 & 0.05437 $\bullet$ & 0.75834 $\bullet$ & 0.71386 $\bullet$\\
&	&	&\\		
SVM & 0.04352 $\,\,\,$ & 0.63836 $\,\,\,$ & 0.78359 $\,\,\,$\\
SVM + BLR + M1 & 0.04583 $\circ$ & 0.62763 $\bullet$ & 0.77509 $\bullet$\\
SVM + BLR + M2 & 0.04327 $\bullet$ & 0.62227 $\bullet$ & 0.78255 $\bullet$\\
SVM + SMO + M1 & 0.04522 $\circ$ & 0.61821 $\bullet$ & 0.76828 $\bullet$\\
SVM + SMO + M2 & 0.04313 $\bullet$  & 0.62135 $\bullet$ & 0.78150
$\bullet$\\\hline
\end{tabular}
\caption{Example-based and ranking measures for Ohsumed-23}
\label{table:ohsumed23_multi}
\end{center}
\end{table}

\begin{table}
\begin{center}
\begin{tabular}{|c|ccc|}\hline
Model	&Hamming loss	& Subset 0/1 loss & One error\\\hline
NB $\,\,\,$& 0.04423 $\,\,\,$ &0.92348 $\,\,\,$& 0.68809 $\,\,\,$\\
NB + BLR + M1 &0.04090 $\bullet$ &0.91627 $\bullet$ &0.69332 $\circ$\\
NB + BLR + M2 &0.04079 $\bullet$ &0.91494 $\bullet$ &0.69433 $\circ$\\
NB + SMO + M1	&0.04486 $\circ$	 &0.93294 $\circ$	&0.53343 $\bullet$\\
NB + SMO + M2	&0.04559 $\circ$	 &0.93699 $\circ$	&0.51662 $\bullet$\\
&	&	&\\	
KNN $\,\,\,$&	0.02263 $\,\,\,$ & 0.70625 $\,\,\,$ &	0.76917 $\,\,\,$\\
KNN + BLR + M1	&0.03624 $\circ$	&0.75544 $\circ$	&0.61414 $\bullet$\\
KNN + BLR + M2	&0.02413 $\circ$	&0.73446 $\circ$	&0.76842 $\bullet$\\
KNN + SMO + M1	&0.04320 $\circ$	&0.88471 $\circ$	&0.48724 $\bullet$\\
KNN + SMO + M2	&0.04176 $\circ$	&0.93213 $\circ$	&0.36426 $\bullet$\\
&	&	&\\					
SVM	&0.01187 $\,\,\,$	&0.51391 $\,\,\,$	&0.95040 $\,\,\,$\\
SVM + BLR + M1	&0.01461 $\circ$	&0.53488 $\circ$	&0.93696 $\bullet$\\
SVM + BLR + M2	&0.01254 $\circ$	&0.50866 $\bullet$	&0.95597 $\circ$\\
SVM + SMO + M1	&0.02414 $\circ$	&0.89497 $\circ$	&0.80030 $\bullet$\\
SVM + SMO + M2	&0.02618 $\circ$	&0.91824 $\circ$	&0.60191 $\bullet$\\\hline
\end{tabular}
\caption{Example-based and ranking measures for RCV1}
\label{table:rcv1_multi}
\end{center}
\end{table}

\begin{table}
 \begin{center}
\begin{tabular}{|c|c|c|c|}\hline
 &Hamming loss	& Subset 0/1 loss & One error\\\hline
BLR + M1 & 3/\textit{6} & 5/\textit{4} & 8/\textit{1} \\
BLR + M2 & 5/\textit{4} & 7/\textit{2} & 7/\textit{2} \\ 
SMO + M1& 2/\textit{7} & 4/\textit{5} & 9/\textit{0} \\
SMO + M2& 3/\textit{6} & 4/\textit{5} & 8/\textit{0} \\\hline
 \end{tabular}
\caption{Number of times that the baseline classifier plus either M1 or M2
(using either BLR or SMO as label classifiers) is better/\textit{worse} than the corresponding
baseline classifier alone, with respect to different example-based and rank
measures}
\label{table:resumen2}
\end{center}
\end{table}

The results show different patterns for each one of the measures. One error
behaves really good,
showing a very good performance in all three collections, improving the baseline
both in the M1 and M2 versions for almost all the cases. For the rest of measures, they tend to work well with naive Bayes
(perhaps the weakest of the three
classifiers). Subset 0/1 loss is improved in all cases in Ohsumed-23 but only in
5 and in 3 cases in Reuters and RCV1, respectively.
It seems that, for most of the cases, M2 is better than M1 for this measure, and
BLR better than SMO.

Hamming loss shows a good performance for Ohsumed-23, where all the M2 versions
of our classifier outperform 
the baselines. In Reuters, however, only 5 out of 12 cases a performance
improvement was shown. This fact is 
also present in RCV1, where only for naive Bayes and BLR an improvement is
obtained. Again, M2 presents a better performance than M1
in general. The fact that the two example-based measures do not show great
improvements (and even decrease their performance)
can be explained with the fact that our method tends to be designed to produce
great increments in {\em macro-} measures (i.e.
per category) and these measures are averaged for each instance (that is, they
are close to the {\em micro-} ones), except
of the fact that Hamming loss accounts for the number of false positives and
negatives, regardless of how many other labels
have been predicted well. A summary of these results is shown in Figure
\ref{table:resumen2}, where the number of
times that each one of our models is better for each measure than the corresponding
baseline is shown.

\section{Conclusions and future works}\label{sec:conclusions}

In this paper we have proposed a quite general methodology to better manage
multilabel classification problems. It is based on explicitly taking into
account the dependences among labels. The proposed method can use as the
starting point any set of binary classifiers (one for each label, which are
trained from the content of the instances in the training set) able to produce a
probabilistic output. This information is merged in a principled way with the
information generated by another set of binary classifiers, which in this case
are trained using only the information about the labels assigned to the
instances in the training set (capturing the dependences among labels). These
label-based classifiers, like the content-based ones, can be built using a
variety of learning methods, provided that their output can be interpreted
probabilistically. 

We have carried out experiments with three well-known multilabel document
collections, using three very different content-based baseline classifiers
(naive Bayes, k-NN and SVM) and two label-based classifiers (logistic
regression and SVM). The experiments confirm that our methodology often tends
to 
improve the results obtained by the baseline classifiers.

The combination of our methods (which so far use a natural threshold of 0.5)
with any of the well-known thresholding techniques \cite{yang2} is an open
question. The first and obvious question is: what happens if a thresholding
algorithm is applied after our method (instead of just binarizing the final
probability output)? At a first glance it should improve the results, but how?
Is this combination worthwhile? Also, a more sophisticated variant of the M1
method, with a better thresholding function $\tau$ (using learned values for the
thresholds, not just $0.5$) can be studied. Or even, a combination of both
proposals. As future work we would like to study the synergies between our
proposal for improving multilabel classification and some thresholding
techniques.

Another open question is the following: using the same approach, a different
independence assumption than the one given in eq. (\ref{nb})
could be given, leading to other models. In particular, we would like to explore
methods where the independence assumption is relaxed, because we think that a
complete independence may be unrealistic in some categories.

Finally, we would like to use this methodology in a different environment. We
have selected document categorization for validating this model because it is a
natural form of multilabeled instances. In the future, we would like to work
with different datasets as, for example, musical patterns, protein data or
social network data, which we think are also suitable for this
method.\vspace{.5cm} 

\noindent {\bf Acknowledgements}
This study has been jointly supported by the Spanish research programme
Consolider Ingenio 2010 and the Consejer\'{\i}a de Innovaci\'on, Ciencia y
Empresa de la Junta de Andaluc\'{\i}a, under the projects MIPRCV (CSD2007-00018)
and P09-TIC-4526, respectively.

\bibliographystyle{plain}
\bibliography{bibliography}

\begin{thebibliography}{10}

\bibitem{libsvm}
Chih-Chung Chang and Chih-Jen Lin.
\newblock {LIBSVM}: A library for support vector machines.
\newblock {\em ACM T. Intel. Syst. Tech.}, 2:27:1--27:27, 2011.
\newblock Software available at \url{http://www.csie.ntu.edu.tw/~cjlin/libsvm}.

\bibitem{cheng}
Weiwei Cheng and Eyke H\"{u}llermeier.
\newblock Combining instance-based learning and logistic regression for
  multilabel classification.
\newblock {\em Mach. Learn.}, 76(2-3):211--225, September 2009.

\bibitem{clare}
A.~Clare and R.~D. King.
\newblock Knowledge discovery in {Multi-Label} phenotype data.
\newblock In {\em Proceedings of the 5th European Conference on Principles of
  Data Mining and Knowledge Discovery (PKDD 2001)}, pages 42--53, Freiburg,
  Germany, 2001.

\bibitem{ijar09}
Luis~M. de~Campos and Alfonso~E. Romero.
\newblock Bayesian network models for hierarchical text classification from a
  thesaurus.
\newblock {\em Int. J. Approx. Reason.}, 50(7):932--944, July 2009.

\bibitem{dembe}
Krzysztof Dembczynski, Weiwei Cheng, and Eyke H{\"u}llermeier.
\newblock Bayes optimal multilabel classification via probabilistic classifier
  chains.
\newblock In {\em ICML}, pages 279--286, 2010.

\bibitem{dembe2}
Krzysztof Dembczy{\'n}ski, Willem Waegeman, Weiwei Cheng, and Eyke
  H\"ullermeier.
\newblock On label dependence and loss minimization in multi-label
  classification.
\newblock {\em Mach. Learn.}, 88(1-2):5--45, 2012.

\bibitem{dimou}
Anastasios Dimou, Grigorios Tsoumakas, Vasileios Mezaris, Ioannis Kompatsiaris,
  and Ioannis Vlahavas.
\newblock An empirical study of multi-label learning methods for video
  annotation.
\newblock In {\em Proceedings of the 2009 Seventh International Workshop on
  Content-Based Multimedia Indexing}, CBMI '09, pages 19--24, Washington, DC,
  USA, 2009. IEEE Computer Society.

\bibitem{genkin}
Alexander Genkin, David~D. Lewis, and David Madigan.
\newblock Large-scale bayesian logistic regression for text categorization.
\newblock {\em Technometrics}, 49:291--304(14), 2007.

\bibitem{godbole}
Shantanu Godbole and Sunita Sarawagi.
\newblock Discriminative methods for multi-labeled classification.
\newblock {\em Adv. Knowl. Disc. Data Min.}, pages 22--30, 2004.

\bibitem{ohsu}
William Hersh, Chris Buckley, T.~J. Leone, and David Hickam.
\newblock Ohsumed: an interactive retrieval evaluation and new large test
  collection for research.
\newblock In {\em Proceedings of the 17th annual international ACM SIGIR
  conference on Research and development in information retrieval}, SIGIR '94,
  pages 192--201, New York, NY, USA, 1994. Springer-Verlag New York, Inc.

\bibitem{joachimsSVM}
Thorsten Joachims.
\newblock Text categorization with suport vector machines: Learning with many
  relevant features.
\newblock In {\em Proceedings of the 10th European Conference on Machine
  Learning}, ECML '98, pages 137--142, London, UK, UK, 1998. Springer-Verlag.

\bibitem{lewisr}
D.~D. Lewis and M.~Ringuette.
\newblock A comparison of two learning algorithms for text categorization.
\newblock In {\em Symposium on Document Analysis and Information Retrieval.},
  1994.

\bibitem{lewisrcv1}
David~D. Lewis, Yiming Yang, Tony~G. Rose, and Fan Li.
\newblock Rcv1: A new benchmark collection for text categorization research.
\newblock {\em J. Mach. Learn. Res.}, 5:361--397, December 2004.

\bibitem{luo}
Xiao Luo and A.~Nur Zincir-Heywood.
\newblock Evaluation of two systems on multi-class multi-label document
  classification.
\newblock In {\em Proceedings of the 15th international conference on
  Foundations of Intelligent Systems}, ISMIS'05, pages 161--169, Berlin,
  Heidelberg, 2005. Springer-Verlag.

\bibitem{pr12}
Gjorgji Madjarov, Dragi Kocev, Dejan Gjorgjevikj, and Sa\v{s}o D\v{z}eroski.
\newblock An extensive experimental comparison of methods for multi-label
  learning.
\newblock {\em Pattern Recogn.}, 45:3084--3104, 2012.

\bibitem{mccallum}
Andrew McCallum and Kamal Nigam.
\newblock A comparison of event models for naive bayes text classification.
\newblock In {\em AAAI-98 Workshope on Learning for Text Categorization}, pages
  41--48. AAAI Press, 1998.

\bibitem{mccallum99}
Andrew~Kachites McCallum.
\newblock Multi-label text classification with a mixture model trained by em.
\newblock In {\em AAAI 99 Workshop on Text Learning}, 1999.

\bibitem{pandey}
Gaurav Pandey, Vipin Kumar, and Michael Steinbach.
\newblock Computational approaches for protein function prediction: A survey.
\newblock Technical Report 06-028, Department of Computer Science and
  Engineering, University of Minnesota, Twin Cities, 2006.

\bibitem{platt}
J.~Platt.
\newblock Probabilistic outputs for support vector machines and comparison to
  regularized likelihood methods.
\newblock In {\em Advances in Large Margin Classifiers}, 2000.

\bibitem{snowball}
Martin~F. Porter.
\newblock Snowball: A language for stemming algorithms.
\newblock Published online, October 2001.
\newblock Accessed 11.03.2008, 15.00h.

\bibitem{daurolab}
Alfonso~E. Romero.
\newblock Daurolab.
\newblock Published online, 2010.
\newblock Software available at \url{http://sourceforge.net/projects/daurolab}.

\bibitem{Salton3}
G.~Salton and M.~E. Lesk.
\newblock The smart automatic document retrieval systems\--an illustration.
\newblock {\em Commun. ACM}, 8(6):391--398, June 1965.

\bibitem{Sebastiani}
Fabrizio Sebastiani.
\newblock Machine learning in automated text categorization.
\newblock {\em ACM Comput. Surv.}, 34(1):1--47, March 2002.

\bibitem{su11}
Ja-Hwung Su, Chien-Li Chou, Ching-Yung Lin, and V.S. Tseng.
\newblock Effective semantic annotation by image-to-concept distribution model.
\newblock {\em IEEE T. Multimedia}, 13(3):530 --538, june 2011.

\bibitem{taylor10}
Portia~E. Taylor, Gustavo J.~M. Almeida, Takeo Kanade, and Jessica~K. Hodgins.
\newblock Classifying human motion quality for knee osteoarthritis using
  accelerometers.
\newblock {\em Conf. Proc. IEEE Eng. Med. Biol. Soc}, 1:339--43, 2010.

\bibitem{trohidis}
K.~Trohidis, G.~Tsoumakas, G.~Kalliris, and I.~Vlahavas.
\newblock {Multilabel Classification of Music into Emotions}.
\newblock In {\em Proc. 9th International Conference on Music Information
  Retrieval (ISMIR 2008), Philadelphia, PA, USA, 2008}, 2008.

\bibitem{tsoumakas}
Grigorios Tsoumakas and Ioannis Katakis.
\newblock Multi-label classification: An overview.
\newblock {\em IJDWM}, 3(3):1--13, 2007.

\bibitem{miningmulti-label}
Grigorios Tsoumakas, Ioannis Katakis, and Ioannis Vlahavas.
\newblock Mining multi-label data.
\newblock In {\em Data Mining and Knowledge Discovery Handbook}, pages
  667--685, 2010.

\bibitem{ueda}
Naonori Ueda and Kazumi Saito.
\newblock Parametric mixture models for multi-labeled text.
\newblock In {\em Advances in Neural Information Processing Systems 15}, pages
  721--728. MIT Press, 2003.

\bibitem{valentini11}
Giorgio Valentini.
\newblock True path rule hierarchical ensembles for genome-wide gene function
  prediction.
\newblock {\em IEEE/ACM Trans. Comput. Biol. Bioinformatics}, 8(3):832--847,
  May 2011.

\bibitem{vanderwal}
Thomas~Vander Wal.
\newblock Folksonomy coinage and definition.
\newblock Website, Februar 2007.
\newblock \url{http://vanderwal.net/folksonomy.html}.

\bibitem{witten}
Ian~H. Witten and Eibe Frank.
\newblock {\em Data Mining: Practical Machine Learning Tools and Techniques,
  Second Edition (Morgan Kaufmann Series in Data Management Systems)}.
\newblock Morgan Kaufmann Publishers Inc., San Francisco, CA, USA, 2005.

\bibitem{yang11}
Yi-Hsuan Yang and H.H. Chen.
\newblock Ranking-based emotion recognition for music organization and
  retrieval.
\newblock {\em IEEE T. Audio Speech Lang. Proc.}, 19(4):762 --774, may 2011.

\bibitem{yang2}
Yiming Yang.
\newblock A study of thresholding strategies for text categorization.
\newblock In {\em Proceedings of the 24th annual international ACM SIGIR
  conference on Research and development in information retrieval}, SIGIR '01,
  pages 137--145, New York, NY, USA, 2001. ACM.

\bibitem{yang99}
Yiming Yang and Xin Liu.
\newblock A re-examination of text categorization methods.
\newblock In {\em Proceedings of the 22nd annual international ACM SIGIR
  conference on Research and development in information retrieval}, SIGIR '99,
  pages 42--49, New York, NY, USA, 1999. ACM.

\bibitem{zhang}
Min-Ling Zhang and Zhi-Hua Zhou.
\newblock A k-nearest neighbor based algorithm for multi-label classification.
\newblock In {\em Granular Computing, 2005 IEEE International Conference on},
  volume~2, pages 718 -- 721 Vol. 2, july 2005.

\end{thebibliography}

\end{document}